\documentclass{article}

\usepackage{microtype}
\usepackage{graphicx}
\usepackage{subfigure}
\usepackage{booktabs} 

\usepackage{hyperref}

\usepackage[accepted]{icml2024}

\usepackage{amsmath}
\usepackage{amssymb}
\usepackage{mathtools}
\usepackage{amsthm}
\usepackage[inline]{enumitem}

\usepackage[capitalize,noabbrev]{cleveref}

\theoremstyle{plain}

\theoremstyle{definition}

\theoremstyle{remark}

\usepackage[textsize=tiny]{todonotes}

\usepackage{amsmath,amsfonts,bm}

\def\eqref#1{equation~\ref{#1}}

\def\1{\bm{1}}

\def\vtheta{{\bm{\theta}}}

\def\vx{{\bm{x}}}
\def\vy{{\bm{y}}}

\DeclareMathAlphabet{\mathsfit}{\encodingdefault}{\sfdefault}{m}{sl}
\SetMathAlphabet{\mathsfit}{bold}{\encodingdefault}{\sfdefault}{bx}{n}

\newcommand{\E}{\mathbb{E}}

\newcommand{\KL}{D_{\mathrm{KL}}}

\usepackage{multirow}
\usepackage{xspace}

\newcommand{\tool}{OSD\xspace}

\usepackage{multirow}

\usepackage{array}
\usepackage{makecell}
\newcolumntype{P}[1]{>{\centering\arraybackslash}p{#1}}

\icmltitlerunning{Online Speculative Decoding}

\begin{document}

\twocolumn[
\icmltitle{Online Speculative Decoding}



\icmlsetsymbol{equal}{*}

\begin{icmlauthorlist}
\icmlauthor{Xiaoxuan Liu}{ucb}
\icmlauthor{Lanxiang Hu}{ucsd}
\icmlauthor{Peter Bailis}{google}
\icmlauthor{Alvin Cheung}{ucb}
\icmlauthor{Zhijie Deng}{sjtu}
\icmlauthor{Ion Stoica}{ucb}
\icmlauthor{Hao Zhang}{ucsd}
\end{icmlauthorlist}

\icmlaffiliation{ucb}{UC Berkeley}
\icmlaffiliation{ucsd}{UCSD}
\icmlaffiliation{google}{Google Inc.}
\icmlaffiliation{sjtu}{SJTU}

\icmlcorrespondingauthor{Hao, Zhang}{haozhang@ucsd.edu}
\icmlcorrespondingauthor{Zhijie, Deng}{zhijied@sjtu.edu.cn}

\icmlkeywords{Machine Learning, ICML}

\vskip 0.3in
]



\printAffiliationsAndNotice{}  

\begin{abstract}
Speculative decoding is a pivotal technique to accelerate the inference of large language models (LLMs) by employing a smaller draft model to predict the target model's outputs. However, its efficacy can be limited due to the low predictive accuracy of the draft model, particularly when faced with diverse text inputs and a significant capability gap between the draft and target models. 
We introduce online speculative decoding to address this challenge. 
The main idea is to continuously update the (multiple) draft model(s) on observed user query data. 
Adapting to query distribution mitigates the shifts between the training distribution of the draft model and the query distribution, enabling the draft model to more accurately predict the target model's outputs.
We develop a prototype of online speculative decoding based on knowledge distillation and evaluate it using both synthetic and real query data. The results show a substantial increase in the token acceptance rate by 0.1 to 0.65, bringing 1.42$\times$ to 2.17$\times$ latency reduction. Our code is available at \url{https://github.com/LiuXiaoxuanPKU/OSD}.
\end{abstract}

\section{Introduction}

\begin{figure}[h]  
    \centering
    \vspace{-10pt}
    \includegraphics[width=\linewidth]{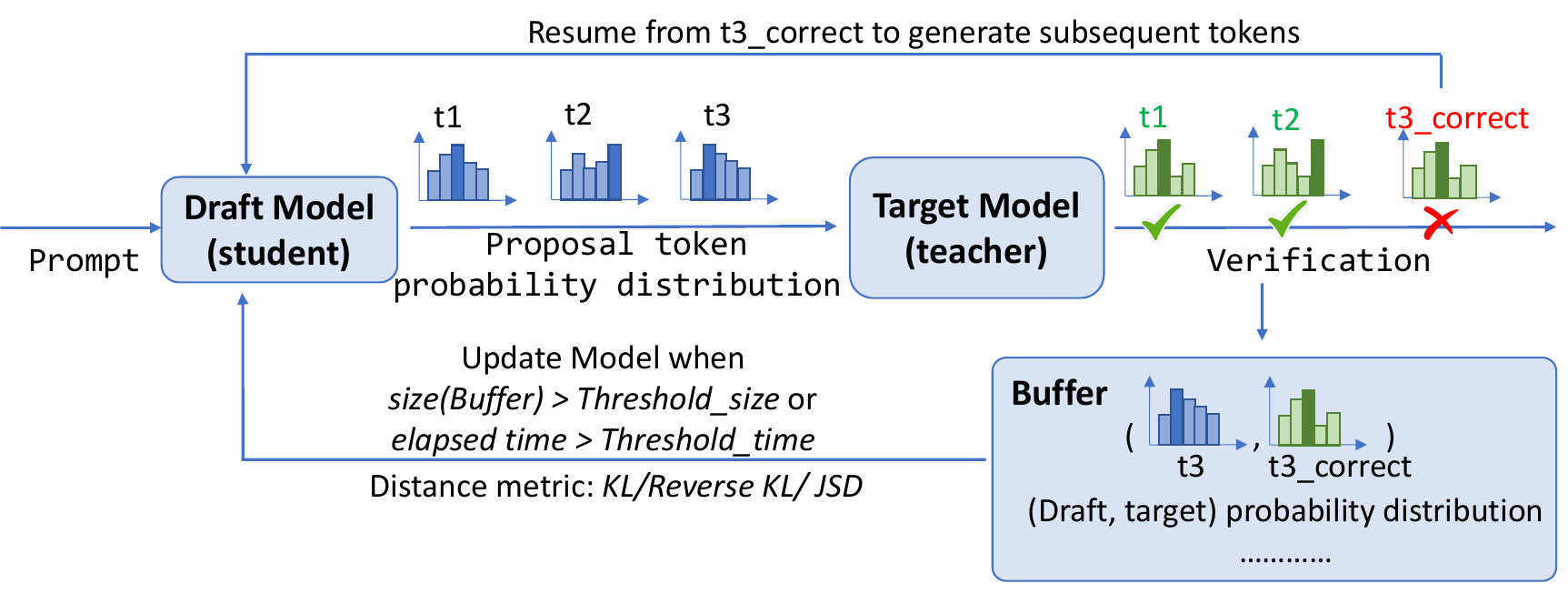}
    \vspace{-20pt}
    \caption{Overview of online speculative decoding (OSD) framework: For each prompt, the draft model suggests multiple tokens and the target model performs the verification. If the student proposes incorrect tokens, both the draft and target distributions are stored in a buffer. Once the buffer exceeds a size limit or is too old, the draft model is updated by calculating the loss between the draft and target distributions using various distance metrics.}
    \vspace{-10pt}
    \label{fig:arch}
\end{figure}

Large language models (LLMs) such as GPT-4~\citep{openai2023gpt} and LLaMA~\citep{touvron2023llama,touvron2023llama2} are rapidly reinventing today's applications. 
Many companies are racing to deploy LLMs in online services, such as search, chatbots, and virtual assistants. Since most of these services demand low latency, optimizing LLM serving latency directly translates into better quality of service and cost reduction.

The latency of today's LLM service is unfortunately very high, primarily because serving a user query requires multiple serial evaluations of the LLM, each generating only one token of the response.
An emerging solution to reduce the latency is speculative decoding~\cite{leviathan2023fast}---it employs a small draft model to speculate multiple output tokens of the target (large) model, and then lets the target LLM verify these speculations in parallel. 
If the verification of a token fails, the large model must recompute from that point. Therefore, the performance of speculative decoding largely depends on the speculation accuracy of the draft model (also known as the token acceptance rate). 
In the presence of diverse text inputs, the accuracy of the speculations is often not very high, due to the capability gap between the draft and target models. 
Employing a larger, more accurate draft model, however, defeats the purpose of speculative decoding as it will increase latency.

To address this challenge, we introduce a novel method, \emph{online speculative decoding} (\tool), to periodically finetune the draft model based on the corrections of the target model. \tool aims to reduce query latency while preserving the compact size of the draft model. 


First, \tool employs knowledge distillation within speculative decoding to enhance the alignment between the draft and target models. 
Speculative decoding involves the draft model proposing potential tokens with their respective probability distributions. 
The target model then assesses these suggestions, correcting discrepancies to ensure that the outputs remain consistent with those produced without the draft model.
This correction mechanism serves as an effective way for the draft model to assimilate and learn from this enriched information. 
Compared to conventional label fine-tuning, knowledge distillation offers a significant advantage by providing a probability distribution for each token. 
By leveraging the insights from the teacher model~\cite{gu2023knowledge}, this method effectively aligns the draft and the target models.


Furthermore, instead of relying on a static draft model, we periodically update the draft model.
This is because user queries to a specific LLM service often exhibit domain-specific distributions~\citep{zheng2023lmsys}, reflecting shared usage patterns.
While accurately speculating the larger model's outputs on \emph{any diverse input} is challenging, it is feasible to enhance the draft model's prediction accuracy,
\emph{only for similar inputs posted to the service}, characterized by the query distribution.
Updates can be implemented through several methods. One approach is to fine-tune the draft model in the background and then apply these updates in real-time after a predetermined period. Alternatively, one can leverage the excess computational capacity of the serving system while it is running, as detailed in\S~\ref{sec:analysis}.
Importantly, the real-time tuning of the draft models enables them to continuously adapt based on incoming query data. This dynamic approach is essential for preserving a high token acceptance rate, ensuring the model remains efficient and current with evolving data and trends.

Lastly, to further improve the token acceptance rate, \tool not only narrows the query distribution but also routes each query to the draft model best suited for that specific distribution. This is accomplished by developing draft models that are finely tuned to 
cater to distinct domains. Concentrating on a narrower query distribution has proven more effective for learning. 
Consequently, \tool efficiently directs queries to the corresponding draft model that specializes in their domain. 
As evidenced in \S~\ref{sec:eval:online_evaluation} of our evaluation, we have adeptly 
trained multiple draft models, each uniquely tailored to distinct languages or topics. This method highlights the significant potential 
for improved efficiency and accuracy when dealing with a diverse range of queries.

In summary, this paper makes the following contributions:
\begin{itemize}[nosep,leftmargin=1em,labelwidth=*,align=left]
    \item We explore various generalized knowledge distillation (GKD) methods for constructing draft models and identify the most effective variants (Section~\ref{sec:knowledge-distill}).
    \item We introduce online speculative decoding to reduce LLM serving latency by adapting draft models on the fly (\S~\ref{sec:adaptation}).
    \item We investigate draft model customization for speculative decoding, wherein each query is directed to the draft model that corresponds with the query's domain (\S~\ref{sec:customization}).
    \item \tool demonstrates a significant improvement in token acceptance rate, by up to 10-65\% on diverse datasets, which translates into a 1.4-2.1$\times$ reduction in latency. 
    \tool can be combined with existing methods that construct static draft models and match the accuracy achieved if all query data were available beforehand (\S~\ref{sec:experiment}).
\end{itemize}

\section{Related Work}

\noindent \textbf{Speculative decoding.}
Speculative decoding~\citep{leviathan2023fast, chen2023accelerating} accelerates LLM decoding by employing a (small) draft model to predict the outputs of the larger target model, which the target model then verifies. 
Suppose the draft model can correctly predict more than one token per verification step, the memory I/O for accessing the weights and KV cache of the (large) target model at inference is amortized across 
multiple output tokens, thereby reducing latency, especially since LLM inference is often constrained by GPU HBM bandwidth.
The efficacy of speculative decoding hinges on the draft model's ability to accurately predict the target model's outputs. Existing work improves the speculation accuracy by using staged~\citep{spector2023accelerating}, RAG-style~\cite{he2023rest}, multiple~\citep{miao2023specinfer, chen2023cascade} draft models and sampling multiple candidates from the draft model~\cite{yang2024multi, cai2024medusa}. 
Additionally, there exists a line of research that eliminates the need for a separate draft model by leveraging auxiliary modules within the target model itself~\citep{medusa, stern2018blockwise, cai2024medusa, lin2024bita, zhang2023draft}. 
These methods predominantly assume a static draft model post-deployment. In contrast, our work introduces a framework that actively adapts the draft model to the evolving user query distribution on the fly, 
irrespective of the draft model's construction. \tool is orthogonal to the aforementioned methods, enabling its integration with them to improve overall efficacy in online deployment scenarios.


\noindent \textbf{Distillation for auto-regressive models.}
Knowledge distillation (KD) is a framework to generate smaller models that emulate the performance of larger models. However, KD in its conventional form has been observed to be less effective for LLMs. 
\citet{gu2023knowledge} extends KD to autoregressive LLMs by decoding from the student model and optimizing the reserve KL divergence between students and teachers.
\citet{agarwal2023gkd} introduce generalized knowledge distillation (GKD) to optimize a linear combination of the forward KL and reverse KL between teacher and student, using a blend of teacher- and student-sampled data. 
Drawing inspiration from both works, \tool applies KD to speculative decoding for LLMs
and extends it to dynamically adjust draft models (Section~\ref{sec:knowledge-distill}).
We acknowledge the simultaneous emergence of a related work, DistillSpec~\cite{zhou2023distillspec}, which also employs KD for speculative decoding. However, our work and DistillSpec were developed concurrently. Moreover, DistillSpec represents a specific aspect of our broader framework \tool. \tool not only explores KD for speculative decoding but also addresses challenges in the online setting and routes queries across various distributions.

\section{Background}

We first briefly review speculative decoding~\citep{leviathan2023fast}, a critical technique that accelerates inference of a large target LLM $p(\cdot|\vx)$ with token proposals from a small draft model $q_\vtheta(\cdot|\vx)$. 
$\vx$ denotes the concatenation of the input prompts and already generated tokens. 
The two distributions are both learned in an auto-regressive way. 
We emphasize the parameters $\vtheta$ of the draft model because we usually need to tailor them according to the target LLM for more substantial acceleration. 

Speculative decoding uses a (small) draft model to propose $k$ tokens ${\vy} \triangleq \{ y_i\}_{i=1}^k \sim q_\vtheta(\cdot | \vx)$, and lets the target LLM estimate the $k+1$ token probabilities $\{p({y}|\vx, {\vy}_{<i})\}_{i=1}^{k+1}$\footnote{${\vy}_{<i}$ refers to $\{ y_j\}_{j=1}^{i-1}$.} in parallel. We detailed the sampling process in Appendix~\ref{appendix:sd}. Prior work has shown that the resulting samples $\tilde{\vy} \triangleq \{{y}_1, \dots, y_{a+1}\}$ strictly follow the distribution of the target LLM $p(\cdot|\vx)$~\citep{leviathan2023fast}. 
It concatenates $\tilde{\vy}$ to $\vx$ and repeats the above process until meeting ⟨EOS⟩. 

\textbf{Expected acceptance rate \& speedups.} 
The acceptance rate, denoted as \(\alpha\), serves as a measure of how closely the draft model approximates the target model. It is defined as the expected probability that speculative decoding will accept a proposal token given the prompt \(y_i \sim q_\vtheta(y_i|\vx, {\vy}_{<i})\). This rate directly influences the expected length 
($\E(|\tilde{\vy}|)$) of \(\tilde{\vy}\) 
for each target LLM run and the speedup of speculative decoding as detailed in Figure~\ref{fig:analysis-alphas}.

Assuming that the \(k + 1\) simultaneous evaluation of the target LLM \(p\) takes roughly the same amount of time as generating a single token in parallel, 
let \(c\) be the time ratio for a single run between \(q_\vtheta\) and \(p\). The expected generation length of a single target LLM run and the speedup in the total wall time due to speculative decoding is represented as~\citep{leviathan2023fast}:
\begin{equation}
\small
\label{eq:gen_len}
    \E(|\tilde{\vy}|) = \frac{1 - \alpha^{k+1}}{1-\alpha},\quad \E(speedup)=\frac{1-\alpha^{k+1}}{(1-\alpha)(kc+1)}.
\end{equation}

{\bf Observation.} The speculative decoding process inherently identifies the inaccuracies of the draft LLM and offers correct solutions for these inaccuracies. Hence, we receive valuable insights on the areas and strategies to refine the draft model at no additional cost. 
Furthermore, given the reduced size of the draft model (e.g., over $20\times$ smaller than the target model), its tuning is not only efficient but also viable for real-time online adjustments. 
Prior work~\citep{leviathan2023fast,miao2023specinfer} has primarily approached speculative decoding in an offline manner, meaning the draft model remains static during online deployment. We next develop online speculative decoding to bridge this gap.

\section{Online Speculative Decoding}
\label{sec:methodology}
We propose the online speculative decoding approach to update the draft model dynamically. 
We frame the learning problem based on the aforementioned auxiliary information as online knowledge distillation, where the teacher and student models correspond to the target and draft LLMs in speculative decoding, respectively.

\subsection{Knowledge Distillation for Speculative Decoding}
\label{sec:knowledge-distill}
Knowledge distillation is a general framework to align the predictive distribution of a small model (i.e., student model) 
with that of a larger one (i.e., teacher model).
Prior research has utilized knowledge distillation to compress neural networks, resulting in decreased inference costs and memory requirements. 
We posit that knowledge distillation is highly effective for speculative decoding. 
In this approach, the draft model acts as the student and the target model serves as the teacher. 
During speculative decoding, we possess complete information on both the proposed and verified probabilities of each token. 
This information helps to construct objectives for distilling the draft model, aligning its output distributions with those of the target model 
and thereby improving the token acceptance rate of the draft model.

The distillation loss generally takes the form of:
\begin{equation}
    \label{eq:distill}
    \small
    \begin{aligned}
        \ell(\vtheta) &= \frac{1}{n_B}\sum_{\vx^{(i)} \in \mathcal{B}} \ell(\vx^{(i)}, \vtheta), \quad \ell(\vx, \vtheta) =  D ({p(\cdot|\vx)} \Vert {q_\vtheta(\cdot|\vx)} ),
    \end{aligned}
\end{equation}
where $\mathcal{B} = \{\vx^{(i)}\}_{i=1}^{n_B}$ denotes a batch of inputs and $D$ denotes some distance measure. 

\textbf{Distance measure.} 
In the case of auto-regressive models, the prediction distribution is categorical at each token. 
Often, we can augment the predicted logits with a tunable temperature $\tau$ for softmax transformation. 
We then use the popular forward KL and reverse KL (RKL), as well as their mixture (i.e., the JSD divergence) 
to instantiate $D$~\citep{agarwal2023gkd,gu2023knowledge}:
\begin{equation}
\small
    \begin{aligned}
       \ell_{KL}(\vx, \vtheta)  &= \KL( {p(\cdot|\vx)}\Vert {q_\vtheta(\cdot|\vx)}), \\
       \ell_{RKL}(\vx, \vtheta)  &= \KL({q_\vtheta(\cdot|\vx)} \Vert {p(\cdot|\vx)}), \\
       \ell_{{JSD}[\beta]} (\vx, \vtheta) &= \beta \KL \left({p(\cdot|\vx)} \big\Vert {p}^\beta_\vtheta(\cdot|\vx)\right) \\
       &\quad + (1-\beta) \KL \left({q_\vtheta(\cdot|\vx)} \big\Vert {p}^\beta_\vtheta(\cdot|\vx)\right),
    \end{aligned}
\end{equation}
where ${p}^\beta_\vtheta(\cdot|\vx) \triangleq \beta{p(\cdot|\vx)} + (1-\beta){q_\vtheta(\cdot|\vx)}$. 
These objectives diverge from the conventionally used label-based fine-tuning objectives in speculative decoding, 
as highlighted in~\citep{miao2023specinfer, leviathan2023fast}. As shown in Section~\ref{sec:offline-eval},
objectives based on the KL divergence prove to be more effective. This is because distributions 
convey richer information than mere labels, thereby enhancing their capability to guide the student model~\citep{hinton2015distilling}. 
Additionally, these objectives enhance convergence rates~\citep{he2022knowledge} and bolster calibration. 
In our study, aligning with previous research~\citep{agarwal2023gkd}, we empirically determine that the optimal 
distance measure can vary depending on the tasks and the relative capacities of the teacher and student models (see \S\ref{sec:offline-eval}).

\textbf{Sampling and gradient estimation.}
Estimating the above objectives involves the expectation over $q_\vtheta(\cdot|\vx)$ or $p(\cdot|\vx)$, which should be expanded recursively. 
Once the recursion depth exceeds $1$, we can not analytically compute $\KL$ but hinge on Monte Carlo approximation. 
When sampling from $q_\vtheta(\cdot|\vx)$, we should differentiate through the sampling process for unbiased gradient estimation. 
However, this leads to policy gradient-style estimators and should rely on elaborate policies such as reward hacking and single-step regularization to reduce gradient variance and stabilize training~\citep{gu2023knowledge}. 

In comparison, a more straightforward approach is to omit the differentiation through the sampling process~\citep{agarwal2023gkd}, where the sample $\vy$ is directly plugged into the objective:
\begin{equation}
\label{eq:offline}
    \small
    \ell(\vx, \vtheta) \approx
 \sum_{j =1}^{|\vy|+1} D({p(y|\vx, \vy_{<j})} \Vert {q_\vtheta(y|\vx, \vy_{<j})} ).
\end{equation}
This way, various distance measures can be readily applied.
Besides, the sampling becomes disentangled from the distance measure. i.e., we sample $\vy$ from an arbitrary mixture of ${p}(\cdot|\vx)$ and ${q}_\theta(\cdot|\vx)$ but use KL, RKL or JSD for estimating the distribution misalignment. 

Intuitively, the samples from the teacher model are usually coherent, which may raise difficulties in fitting the small student model, while samples from the student model may be less structured or even meaningless. 
A workaround strategy is to trade off between them via mixed sampling~\citep{gu2023knowledge}, i.e., $y_j \sim \beta{p(\cdot|\vx, \vy_{<j})} + (1-\beta) q_\vtheta(\cdot|\vx, \vy_{<j})$. 


\subsection{Online Adaptation}
\label{sec:adaptation}
This section expands the application of knowledge distillation for speculative decoding in online environments. 
The approach enables improving the performance of the draft model using results from speculative decoding, 
thus dynamically adapting to the query distribution and improving the token acceptance rate. 
We also discuss the trade-off of our approach when integrating LLM serving systems.

\subsubsection{Algorithm}
\begin{algorithm}[t] 
\caption{Online Speculative Decoding.} 
\label{algo:1}
\small
\begin{algorithmic}[1]
\STATE {\bfseries Input:} 
Target LLM $p(\cdot|\vx)$, draft LLM $q_\vtheta(\cdot|\vx)$, warmup dataset $\mathcal{D}$, online data stream $\mathcal{S}$, guess number $k$, temporary buffer $\mathcal{R}$, replay buffer $\mathcal{Q}$, update interval for the draft model $I$.
\STATE{Pre-train $q_\vtheta$ to approximate $p$ with data from $\mathcal{D}$ by minimizing $\ell(\vx, \vtheta)$ using \cref{eq:offline};}
\STATE{$i \leftarrow 0$;}
\STATE{$\mathcal{Q} \leftarrow []$;}
\STATE{$cur\_len = |\vx|$ // Total sequence length, including prompt length and tokens generated so far.}
\WHILE{True}
    \STATE{$\mathcal{R} \leftarrow []$ // List of ($error\_index$, target logits at $error\_index$) pairs for a single request.}
    \STATE{$\vx \sim \mathcal{S}$, $i \leftarrow i + 1$;}
    \WHILE{⟨EOS⟩ not in $\vx$}
    \STATE{${\vy} = \{{y}_1,...,{y}_k\} \sim q_\vtheta(\cdot|\vx)$;}
    \STATE{Estimate $\{p({y}|\vx, {\vy}_{<i})\}_{i=1}^{k+1}$ in parallel;}
    \STATE{Determine number of accepted tokens $a$ and sample one more token, yielding $\vy=\{{y}_1, \dots, y_{a+1}\}$;} 
    \STATE{$cur\_len \leftarrow cur\_len + a + 1$;}
    \STATE{$error\_index \leftarrow cur\_len$;}
    \STATE{Append $(error\_index, p(y|\vx, {\vy}_{<a+1}))$ to $\mathcal{R}$;}
    \STATE{$\vx \leftarrow [\vx, {\vy}_{<a+2}]$;}
    \ENDWHILE
    \STATE{Append $(\vx, \mathcal{R})$ to $\mathcal{Q}$;}
    \IF{$i\;\mathrm{mod}\;I = 0$}
    \STATE{Update $q_\vtheta$ on $\mathcal{Q}$ to minimize $\ell(\vx, \vtheta)$ analytically;}
    \STATE{$\mathcal{Q} \leftarrow []$;}
    \ENDIF
\ENDWHILE
\end{algorithmic}
\end{algorithm}

We depict our online speculative decoding algorithm (\tool) in \cref{algo:1}.
\tool begins by training the draft model using the warmup dataset (Line 2). 
The serving system then continuously handles incoming requests (as described in Lines 6 to 23).
For each request, it uses standard speculative decoding (Lines 10-11) to generate responses until the ⟨EOS⟩ token. 
Concurrently, \tool tracks the token index ($error\_index$) and target logits where the draft model proposes the wrong tokens (Line 15). 
Leveraging tracked information, \tool updates the draft model every $I$ iteration, with $I$ being a dynamically adjustable parameter.
\tool updates the draft model with different loss functions (Line 20) as described in Section~\ref{sec:knowledge-distill}.
The choice of loss function depends on the specific (draft, target) model pairs and the corresponding input data. 

{\bf Discussion.} 
\tool utilizes a replay buffer, $\mathcal{Q}$, to capture all pertinent information for updating the draft model. 
Various eviction policies can be employed to maintain a compact size for $\mathcal{Q}$. 
For example, one could opt to retain only the most informative pairs or the most recent entries. 
Similarly, users have the option to retain data in $\mathcal{Q}$
even after utilizing it to update the model multiple times.
Determining the optimal eviction/retention strategy is a subject for future exploration. 
In the current study, we refrain from evicting any pairs and release $\mathcal{Q}$ after each model update.
Furthermore, $I$ is a dynamic parameter. Depending on the system load and the rate at which the query
distribution changes, users can adjust $I$ accordingly.
For example, we can perform a gradient update opportunistically only when the service traffic is not on spike (i.e., spare flops are available).

In the implementation of the system, two distinct pipelines can be established: one for training and another for inference. 
This approach allows for the utilization of existing infrastructure. 
Periodic updates of the draft model weights are essential to ensure continuous adaptation. Alternatively, a unified pipeline that accommodates both training and inference can be developed. This integrated system continuously reinforces the draft model, maintaining a consistently high token acceptance rate.
Overall, \tool continuously improves the draft model's approximation (indicated by increased token acceptance rate $\alpha$) 
by learning from the target model during the serving phase. We next demonstrate how the enhanced acceptance rate directly 
contributes to a reduction in request latency.

\subsubsection{Latency \& Flops Analysis} 
\label{sec:analysis}
{\bf Latency.} As detailed in  Appendix~\ref{appendix:latency-analysis}, compared with standard speculative decoding, 
the expected speedup for online speculative decoding is  \( \frac{1+\alpha_2+\alpha_2^2+...+\alpha_2^{k}}{1+\alpha_1+\alpha_1^2+...+\alpha_1^k}\).
Based on the data from our experiment (refer to Table~\ref{tab:apha}), when compared to standard speculative decoding, 
we expect a speedup
improvement for Vicuna-7B (LLaMA-160M as the draft model) by factors of \(2.42\times\), \(1.43\times\), \(1.64\times\), and \(1.22\times\). 
Similarly, for Flan-T5-XL 3B (T5-small 80M as the draft model), the speedup enhancements are \(3.06\times\), \(1.76\times\), \(2.72\times\), and \(1.55\times\) 
across the four evaluated datasets.

{\bf FLOPs.}
(1) The FLOPs required to update the draft model are much fewer than those needed for inference on a large model. 
As elaborated in Appendix~\ref{appendix:flops}, for the two evaluated pairs, 
the FLOPs ratio between the target model and the draft model is 18.75 for the pair (LLaMA-160M, Vicuna7B), and 12.6 for the pair (T5-small 80M, Flan-T5-XL 3B).
(2) In practical systems, the FLOPs required for inference are significantly below the machine's capacity.
Appendix~\ref{appendix:flops} provides an analysis of Arena chatbot traces where the cluster's computational 
utilization is under 1 percent.
Given the above two observations, it becomes evident that the FLOPs spent on inference and updating the draft model are relatively 
insignificant compared with the FLOPs consumed for target model inference and the cluster's total FLOPs.

{\bf Memory Bandwidth.} As detailed in Tabel~\ref{tab:bw} and Appendix~\ref{appendix:bandwidth}, updating the draft model is not memory-intensive because it is small. The majority of memory operations are still dominated by loading the target model. 
OSD can significantly reduce memory bandwidth through a higher token acceptance rate, which consequently decreases the frequency of calling the larger model for verification.


\begin{figure}[ht!]  
    \centering
    \includegraphics[width=0.4\linewidth]{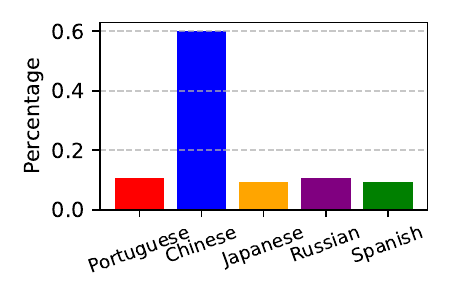}
    \includegraphics[width=0.4\linewidth]{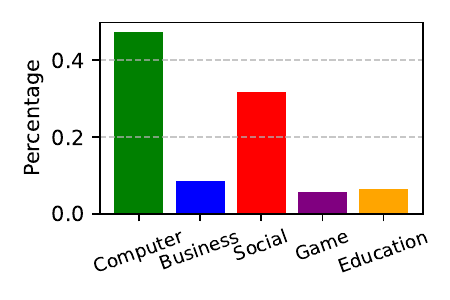}
    \vspace{-10pt}
    \caption{The left plot depicts the proportion of queries in languages other than English in the Arena dataset (detailed in Appendix~\ref{appendix:arena}). The right plot illustrates the distribution of queries across various topics within the dataset.}
    \vspace{-10pt}
    \label{fig:topic-classification}
\end{figure}

\subsection{Draft Model Customization and Routing}
\label{sec:customization}
Queries can be categorized according to various characteristics. Figure~\ref{fig:topic-classification} illustrates two ways to classify queries from the Arena dataset (detailed in Appendix~\ref{appendix:arena}). 
We leverage this observation to further narrow down the query domain to improve the draft model adaptation.
As highlighted in Section~\ref{sec:exp-arena}, we can improve the accuracy of the draft model if we constrain it to a particular domain. This echoes the observation that using a single draft 
model for all queries results in a lower token acceptance rate compared to tailoring the draft model to specific query domains.
Furthermore, we observe a temporal shift in the distribution, detailed in Appendix~\ref{appendix:arena}, 
the proportion of various topics and languages also varies across different timestamps.

In our study, we implemented two approaches for query classification. 
The first approach involves sorting queries based on language, utilizing the language tags available in the query metadata. 
The second strategy clusters queries according to their topics, making use of a small BERT model (67M)~\cite{distill-bert-topic}. 
To minimize routing overhead, we perform routing at the query level rather than at the token level. The routing process is highly efficient as it only relies on either the use of query tags (to access metadata) or the implementation of 
a simple classification model. 
This paper highlights the enhancement in token acceptance rate achieved via the customization of draft models. We leave it as feature work to identify the optimal routing queries and number or size of these customized draft models.
\section{Experiments}
\label{sec:experiment}
To assess the efficacy of \tool, we evaluate its ability to improve the token acceptance rate ($\alpha$) 
within an offline context. 
Subsequently, we examine the approach's impact in an online environment, discovering that the acceptance rate improves even with a moderate amount of data while maintaining 
accuracy levels comparable to those in the offline scenario. Lastly, we investigate query latency and perform an in-depth quantitative analysis to gain insights into the tokens learned by the draft model.

Throughout our experiments, we employ two target models ($M_p)$: Vicuna-7B~\citep{vicuna2023} and FLAN-T5-XL (3B)~\citep{chung2022scaling}. Specifically for Vicuna-7B, we utilize LLaMA-160M~\citep{miao2023specinfer} as the draft model ($M_q$). For FLAN-T5-XL, we use T5-Small~\citep{raffel2020exploring} as the draft model. 
We evaluate performance across four diverse datasets: Text-to-SQL (Spider)~\citep{yu2018spider}, graduate school math (Gsm8k)~\citep{cobbe2021gsm8k}, Python code generation (Code-search-Python)~\citep{husain2019codesearchnet}, and financial question answering (Alpaca-finance)~\citep{alpaca-finance}. 
In all experiments, we set the number of proposed tokens to 5 for speculative decoding. For all online experiments, we fix the update interval \( I \) at 8. 

\subsection{Offline Evaluation}
\label{sec:offline-eval}
In this section, we assess the efficacy of employing knowledge distillation to train a small model specifically for speculation in an offline environment. 
In such a setting, the speculative $M_q$ model has unrestricted access to the dataset, and the query distribution remains stable. 
To emulate these offline conditions, we distill the $M_q$ using the training dataset for two epochs and subsequently evaluate its performance by measuring 
the average token acceptance rate ($\alpha$) on the test set.
As detailed in Section~\ref{sec:knowledge-distill}, we evaluated various sampling methods, namely teacher sampling, student sampling, and mix token-level sampling. 
Table~\ref{tab:apha} displays the token acceptance rate of the draft model for each method, using forward KL as the distance metric on the test dataset. 
For comparison, we also provide the acceptance rate for teacher-generated label fine-tuning and the original model.

For both the Vicuna-7B and FLAN-T5-XL models, the teacher sampling method outperforms others by achieving the highest acceptance rate. 
Furthermore, knowledge distillation has proven its efficacy in enhancing the draft model's approximation, resulting in a high token acceptance rate. 
Lastly, we experimented with different distance measurements like reverse KL and JSD. 
Nevertheless, these measurements 
either paralleled or underperformed when compared to forward KL. 
The optimal distance measurement or 
sampling method varies depending on the task and model, and we leave it to future work to find the best combination. 


\begin{table}
\caption{Token acceptance rates ($\alpha$) after two epochs. 
{\bf FT}: Finetuning on teacher-generated labels. 
{\bf TF, SF, MixF}: Teacher, student, and mix token sampling respectively, all with forward KL.
For tasks, SP: Spider. GS: Gsm8k. CP: Code-serarch-Python. AL: Alpaca-finance.
}
\label{tab:apha}
\begin{center}
\resizebox{0.9\columnwidth}{!}{
\begin{tabular}{ccccccc}
\toprule
{\bf Model}                 & {\bf Task}         &{\bf Original} & {\bf FT} & {\bf TF }   & {\bf SF }    & {\bf MixF}\\
\midrule
\multirow{4}{*}{Vicuna-7B}  & SP             &  0.28    & 0.74     & {\bf 0.76}         & 0.62         & 0.70  \\
                            & GS              &  0.58    & 0.74     & {\bf 0.75}         & 0.67         & 0.73  \\
                            & CP &  0.38    & {\bf 0.65}     & {\bf 0.65}         & 0.51         & 0.61  \\
                            & AL     &  0.57    & {\bf 0.68}     & 0.67         & 0.63         & 0.65  \\
\hline
\multirow{4}{*}{FLAN T5-XL} & SP  &    0.13  &  0.33  & \textbf{0.78}         &      0.67    &  0.70 \\
                            & GS              &  0.29    &  0.50    & \textbf{0.62}         &   0.51      & 0.55  \\
                            & CP &  0.28   &  0.44   & \textbf{0.81}         &    0.67      & 0.78 \\              
                            & AL     &   0.39   &  0.56   & \textbf{0.63}         &    0.59      & 0.60  \\
\bottomrule
\end{tabular}
}
\end{center}
\vspace{-20pt}
\end{table}

\subsection{Online Evaluation}
\label{sec:eval:online_evaluation}
{\bf Online Learning.} First, we evaluate the effectiveness of \tool by addressing two questions: (1) Does the online algorithm increase the token acceptance rate? And is this enhancement comparable to the rates achieved in offline settings, which serve as an upper bound given their full access to data? (2) How quickly does the online algorithm increase the token acceptance rate, thereby indicating that the compact model has grasped the underlying distribution?

In our approach, we replicate the online serving process by iterating through the datasets, extracting prompts, and streaming generation requests. The system utilizes speculative decoding for each of these requests. Throughout this serving phase, we continually refine the speculative models, as detailed in Algorithm~\ref{algo:1}.
For our baseline, we envision a scenario where the serving system has the capability to collect data offline in order to distill an initial draft model. This model is subsequently deployed online to cater to future requests. This process is simulated by using 10\% of the dataset to distill the draft model, which remains static during online serving.
For evaluation metrics, we calculate token acceptance rates averaged over the most recent 50 requests. This demonstrates $M_q$'s efficacy on the most current data.

\begin{figure*}[h]      
    \centering
    \includegraphics[width=0.5\linewidth]{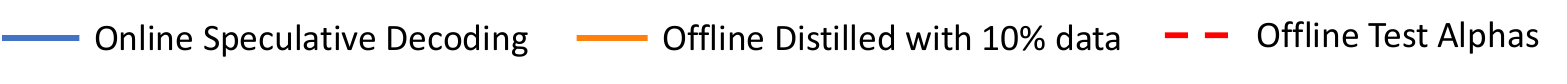}\\
    \includegraphics[width=0.21\linewidth]{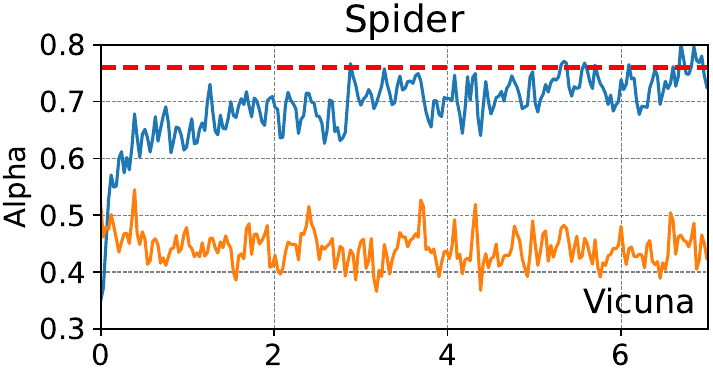}
    \includegraphics[width=0.19\linewidth]{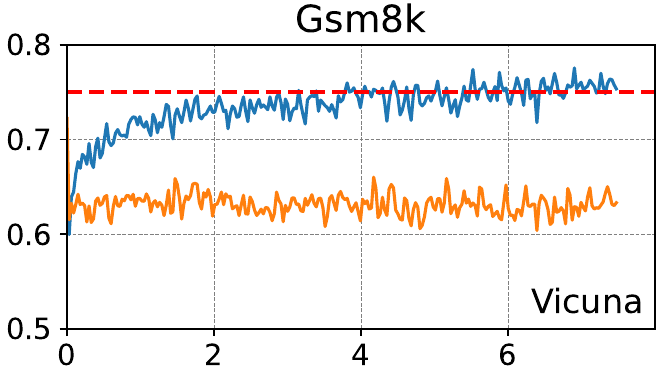}
    \includegraphics[width=0.19\linewidth]{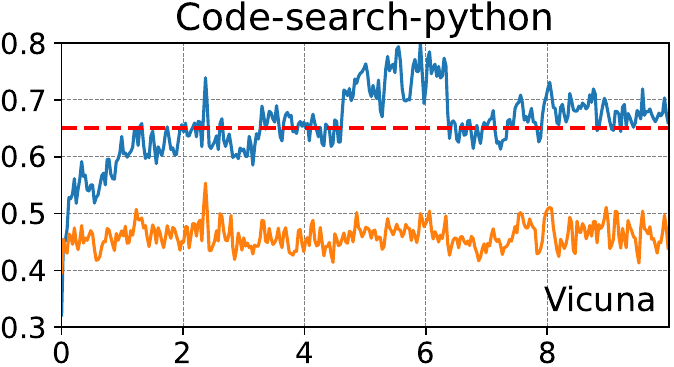}
    \includegraphics[width=0.19\linewidth]{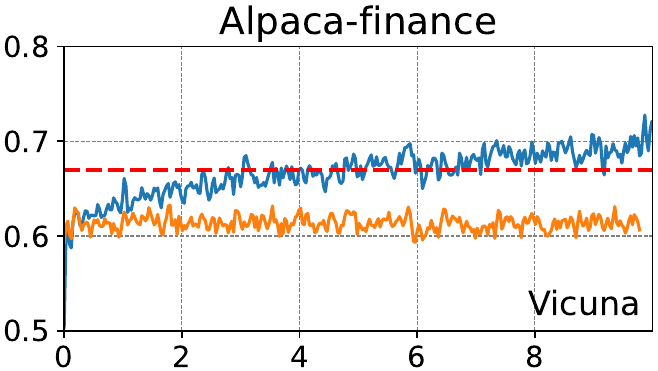}
    \includegraphics[width=0.21\linewidth]{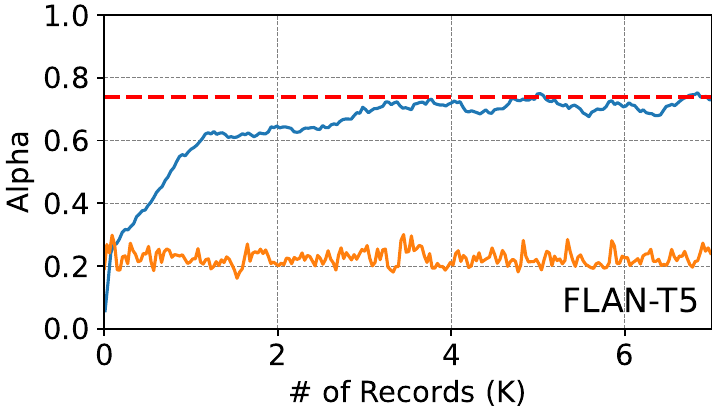}
    \includegraphics[width=0.19\linewidth]{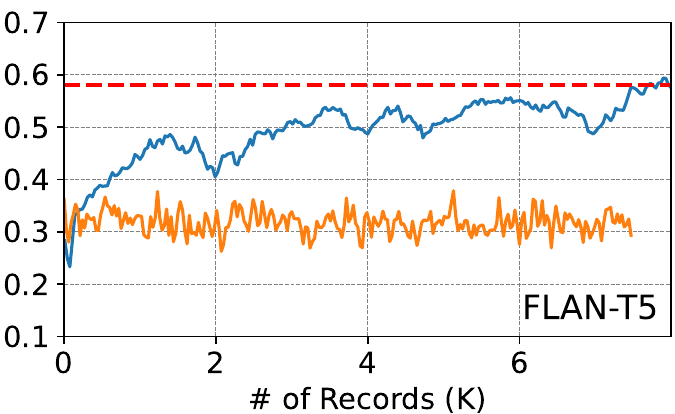}
    \includegraphics[width=0.19\linewidth]{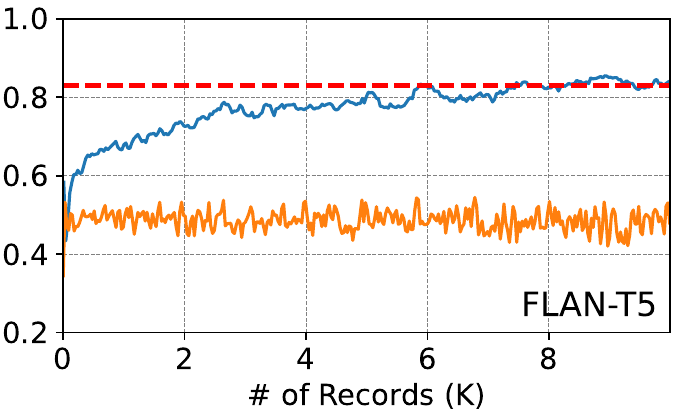}
    \includegraphics[width=0.19\linewidth]{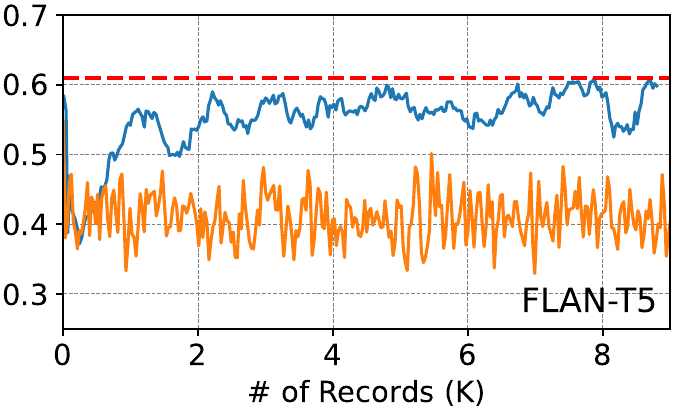}
    \vspace{-5pt}
    \caption{Online acceptance rate ($\alpha$) across different datasets. The x-axis represents the number of records that \tool has processed. $\alpha$ is averaged over the most recent 50 records.}
    \label{fig:alphas}
\end{figure*}

\begin{figure}[h]  
    \centering
    \vspace{-10pt}
    \includegraphics[width=0.95\linewidth]{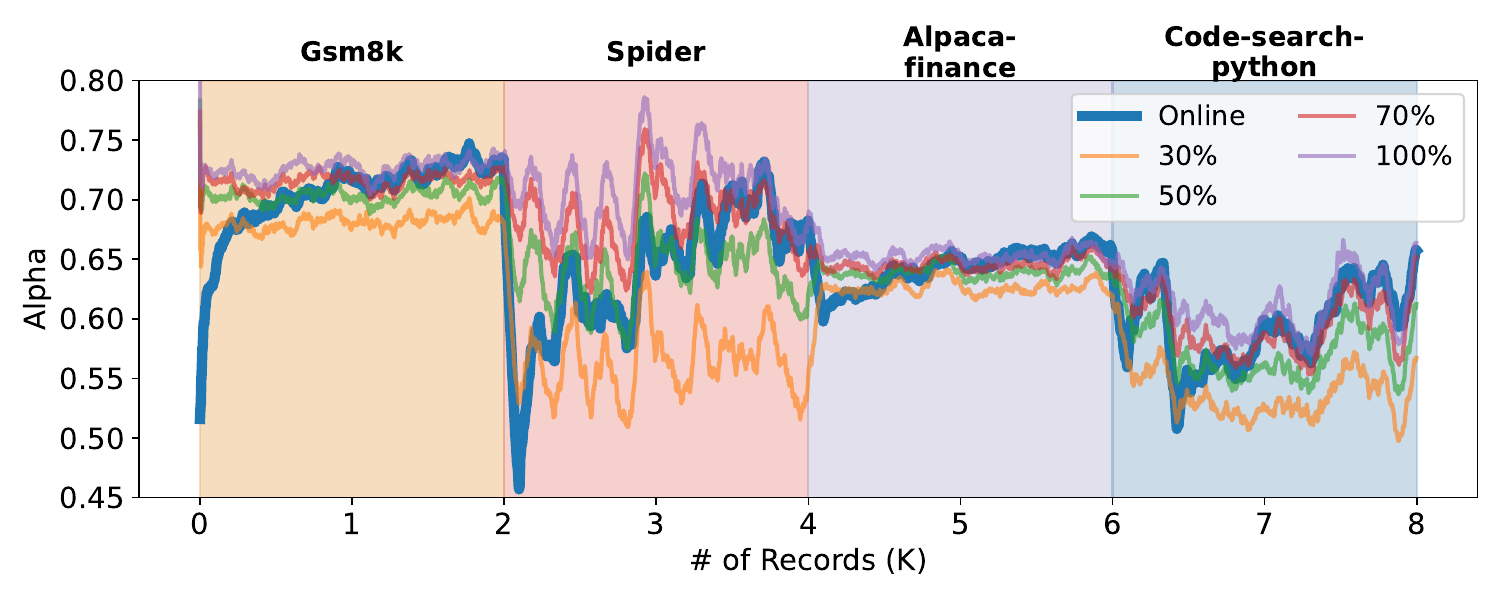}
    \vspace{-10pt}
    \caption{Distribution Shift}
    \vspace{-20pt}
    \label{fig:dis-shift}
\end{figure}

As depicted in Figure 2, both for Vicuna-7B and FLAN-T5, in the beginning, \tool yields a lower token acceptance rate in comparison to the offline distilled model. Nevertheless, these acceptance rates rise swiftly as the draft model is exposed to more data. We also annotate the token acceptance rate from the offline setting to highlight the potential peak performance that the online serving system could reach. 
In all instances, the online context can achieve comparable results. In some scenarios, \tool even surpasses the token acceptance rate of the offline test alphas. This discrepancy can be attributed to the fact that offline test alphas are assessed on the entire test dataset, whereas the online alphas represent the moving average of the latest 50 requests. It's plausible that \tool performs optimally on specific data subsets, particularly if those subsets are more narrowly distributed than the complete dataset.

{\bf Distribution Shifts.} 
We evaluate \tool's ability to adapt to distribution shifts. 
We employ a single LLaMA-160M as the initial draft model and Vicuna-7B as the target model. To simulate the distribution shift, we integrate data from diverse datasets. Concretely, we select 2k prompts from each dataset. The data from the four datasets are amalgamated by direct concatenation, 
such that the records from $i\times2k$ to $(i+1)\times2k$ belong solely to dataset 
$i$. 

As illustrated in Figure~\ref{fig:dis-shift}, \tool's alpha value dips notably at distribution boundaries, especially around 2K, 4K, and 6K records. 
This is anticipated since the draft model initially struggles when faced with a new distribution. However, the alpha value rebounds quickly as \tool processes more data, highlighting its adaptability to shifting query distributions.

We also compared our results to those from a static setting. To ensure the draft model wasn't just memorizing data, we chose samples distinct from the online evaluation data. These samples correspond to 30\%, 50\%, 70\%, and 100\% of each dataset's online evaluation volume, at 0.6K, 1K, 1.4K, and 2K quantities respectively. As depicted in Figure~\ref{fig:dis-shift}, upon an initial shift in query distribution, \tool's performance aligns with or slightly trails the distillation with 30\% data. However, it quickly catches up, matching or even surpassing performances seen with 70\% to 100\% data access. This highlights \tool's ability to rival models fully exposed to the query distribution, even without intimate knowledge of the underlying query dynamics.


\subsection{Real Workloads \& Customized Draft Model} 
\label{sec:exp-arena}
We evaluate \tool on real LMSYS-chat conversations (Appendix ~\ref{appendix:arena}) that span 4 months.
We propose that employing distinct draft models for queries on various subcategories can enhance the token acceptance rate.
To carry out the experiment, we compare two modes: \emph{Single Draft Model} and \emph{Separate Draft Models}. 
Under the \emph{Single Draft Model} case, the target model is paired with a singular draft model through which all data is processed uniformly. Conversely, in the \emph{Separate Draft Models} case, while the target model remains unchanged, it is supported by multiple draft models. Here, each draft model is specialized to handle a specific topic. Upon receiving a new request, the query is first classified using some classification models to determine its relevant topic. Subsequently, the query is directed to the appropriate draft model, tailored for that topic, which then speculates and forwards the requests to the target model. 
Unlike the approach of utilizing a single draft model for all topics, assigning specific draft models to individual topics narrows the range of query distributions each model must adapt to. This focused approach simplifies the learning process for each draft model, as they deal with a more limited set of queries. 

\begin{figure*}[ht!]
    \centering
    \vspace{-5pt}
    \includegraphics[width=0.3\linewidth]{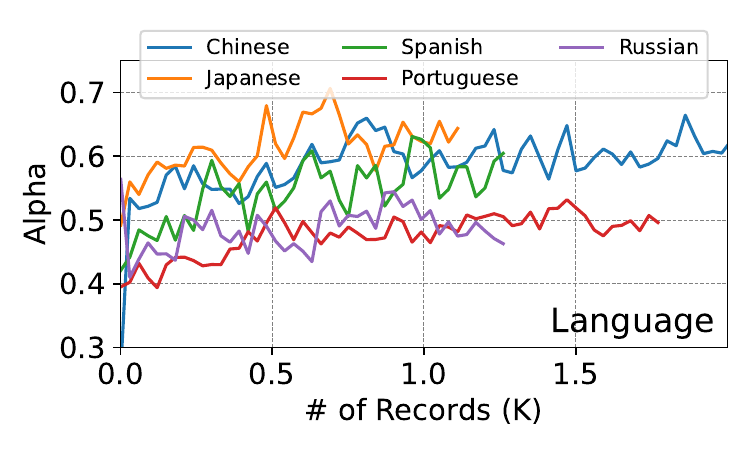}
    \includegraphics[width=0.3\linewidth]{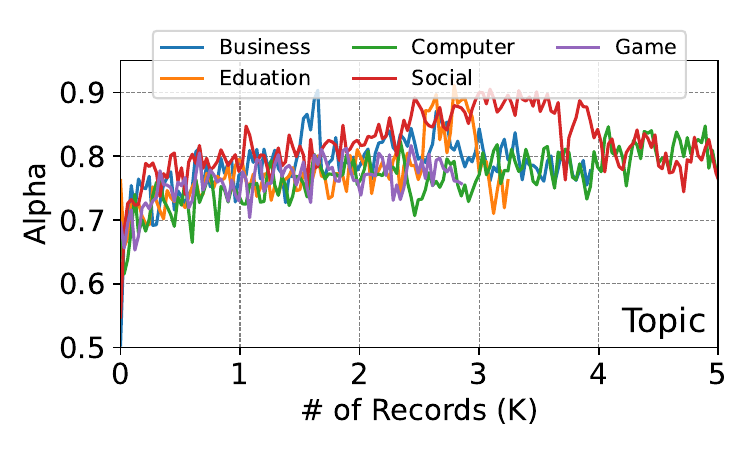}
    \vspace{-10pt}
    \caption{Chatbot Arena Conversations clustered by language and topic.}
    \vspace{-10pt}
    \label{fig:arena}
\end{figure*}

First, we categorize conversations based on the language and we focus on conversations among the top five languages, excluding English. For every chosen language, we use an independent LLaMA-160M to serve as our draft model. All draft models share the same Vicuna-7B as the target model. The token acceptance rate, averaged over the latest 100 requests, showed in Figure~\ref{fig:arena}, reveals that \tool's enhances rates by 0.1 to 0.2, even with under 2K data points. Notably, Japanese was the easiest while Portuguese was the toughest.

Next, we try another way of clustering conversations by routing conversations to draft models with specified for different topics. We get the queries' topics using the fine-tuned distilled Bert model~\citep{distill-bert-topic}, focusing on the top five. For topics with over 5K conversations, we sampled evenly to keep it within 5K. Figure~\ref{fig:arena} shows acceptance rates above 0.6 across topics, with Social and Computer discussions peaking near 0.9. 

To further compare the \emph{Single Draft Model} and \emph{Separate Draft Models} cases, we analyze the token acceptance rate and memory consumption below.

\begin{figure*}[ht!]
    \centering
    \includegraphics[width=0.98\linewidth]{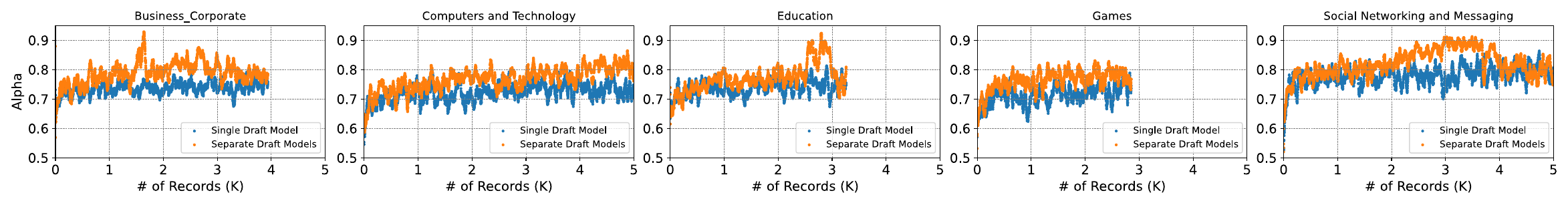}
    \vspace{-10pt}
    \caption{For the single draft model, we send all queries to the same draft model and measure the token acceptance rate based on query topics. For multiple draft models, we employ a customized draft model for each query based on the topic.}
    \label{fig:arena-single-multi}
\end{figure*}

\textbf{Token acceptance rate:} we measured and plotted the token acceptance rates using - a single universal draft model versus multiple topic-specific draft models - in \autoref{fig:arena-single-multi}, to highlight the idea of customizing draft models for different types of queries. As seen from the graph, across all topics, employing multiple draft models results in an increase in the token acceptance rate by 0.1 to 0.2. This aligns with our expectation that draft models benefit from a narrower query distribution, making it easier to learn and adapt. 

\textbf{Memory consumption:} In the analysis below, we assume that multiple draft models are preloaded into memory. When considering five draft models, the cumulative size reaches 800M, approximately 10\% of the target model's size (7B), both in terms of model weights and key-value (kv) cache size. Compared with using a single draft model, routing increases the memory overhead from ~2\% to ~10\%. However, relative to the size of the target model, this additional memory requirement should be manageable. We do think there is the trade-off between the number of draft models and the memory consumption and leave it to future research to decide the optimal number of draft models and the best classification strategy.

\begin{figure}[ht!]
    \vspace{-15pt}
    \centering
    \includegraphics[width=0.49\linewidth]{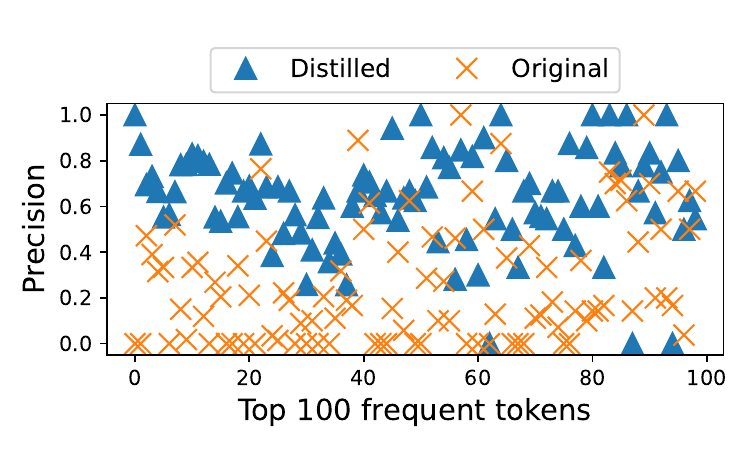}
    \includegraphics[width=0.49\linewidth]{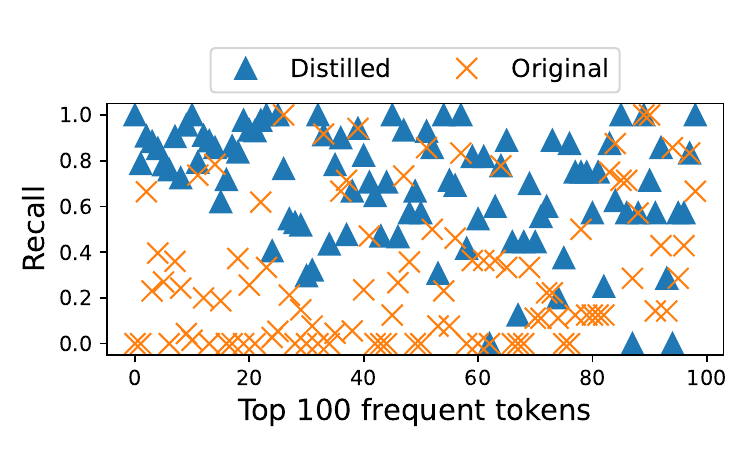}
    \vspace{-10pt}
    \caption{Precision and recall of high-frequency tokens. The x-axis shows token ratings based on occurrence. For instance, token 1 appears most frequently in answers. Precision = \#  of times token $i$ is accepted / \# of times token $i$ is proposed. Recall = \# of times token $i$ is accepted / \# of times token $i$ appears in the final answer.}
    \label{fig:freq-acc}
\end{figure}

\subsection{Latency Measurement}
In this section, we measure the \tool latency at various acceptance rates. 
Subsequently, we evaluate the performance enhancement offered by distilled models across four evaluated datasets. 
Throughout these experiments, we investigate two configurations: (LLaMA-160M draft, Vicuna-7B target) and (TinyLLaMA-1.1B draft, Vicuna-33B target). We conduct the experiments with llamacpp~\cite{llamacpp} on a single A100-80G.

As illustrated in Table~\ref{tab:latency1}, for an identical token acceptance rate, the combination of TinyLLaMA-1.1B and Vicuna-33B outperforms the LLaMA-160M and Vicuna-7B pair in terms of speedup. This enhanced performance is attributable to a greater latency difference between the draft and target models in the first pair (denoted by $c=0.08$ for TinyLLaMA-1.1B and Vicuna-33B, compared to $c=0.13$ for LLaMA-160M and Vicuna-7B). In simpler terms, the draft model incurs relatively lower costs, leading to a more pronounced speedup with equivalent speculation accuracy. Overall, \tool can achieve a maximum speedup of 2.63$\times$ when the token acceptance rate exceeds 0.9. 
Lastly, we evaluate the speedup achieved by \tool's distilled model on four evaluated datasets. For the draft model, as discussed in Section~\ref{sec:knowledge-distill}, we employ teacher sampling with forward KD as the distillation method, as detailed in Table~\ref{tab:latency2}. The results indicate that our \tool can yield a speedup ranging from 1.42$\times$ to 2.17$\times$ across the four evaluated datasets.

\begin{table}[t]
\vspace{-10pt}
\caption{Measured itern token latency (ms) and speedup across $\alpha$s  measured on a single A100-80G with batch size 1. Original is inference without speculative decoding.}
\centering
    \resizebox{0.99\linewidth}{!}{
    \begin{tabular}{P{2.5cm}|P{1.4cm}|P{1.4cm}|P{1.4cm}|P{1.4cm}|P{1.4cm}|P{1.4cm}}
    \toprule
     & Original &  \text{OSD}, $\alpha=0.5$ & \text{OSD}, $\alpha=0.6$ &  \text{OSD}, $\alpha=0.7$ & \text{OSD}, $\alpha=0.8$ & \text{OSD}, $\alpha=0.9$ \\
      \midrule
     \makecell{(1.1B, 33B)  \\ Time (speedup)} & 51.09 & \makecell{39.90 \\(1.28 $\times$)} & \makecell{35.48  \\(1.44 $\times$)} & \makecell{30.96 \\ (1.65 $\times$)} & \makecell{25.42 \\ (2.01 $\times$)} & \makecell{19.43 \\ (2.63 $\times$)} \\
      \hline
     \makecell{(160M, 7B) \\ Time (speedup)} & 13.21  & \makecell{13.85 \\(0.95$\times$)} & \makecell{11.39 \\ (1.17$\times$)} & \makecell{10.20 \\ (1.3$\times$)}  & \makecell{ 7.84 \\($1.68\times$)} & \makecell{5.17 \\ (2.55$\times$)} \\
    \bottomrule
    \end{tabular}
    }
    \label{tab:latency1}
\end{table}

\begin{table}
\vspace{-10pt}
\caption{Measured intern token latency (ms) and speedup on four datasets. 1.1B: TinyLLaMA-1.1B.  33B: Vicuna-33B. 160M: Vicuna-160M. 7B: Vicuna-7B.
\vspace{-10pt}
}
\small
\begin{center}
    \resizebox{1\linewidth}{!}{
    \begin{tabular}{P{3cm}|P{1.4cm}|P{1.4cm}|P{1.4cm}|P{1.4cm}|P{1.4cm}}
    \toprule
    Dataset &  Spider & Gsm8k &  Alpaca-Finance & Code-Python & Chatbot-Arena\\
    \midrule
    \makecell{(1.1B, 33B) \\ Time (Speedup)} & \makecell{23.53 \\ (2.17 $\times$)} & \makecell{27.40 \\(1.89 $\times$)} & \makecell{26.53 \\ (1.92 $\times$)} & \makecell{30.12 \\ (1.69 $\times$)} & \makecell{33.52 \\ (1.51$\times$)} \\
     \hline
    \makecell{(160M, 7B) \\ Time (Speedup)} & \makecell{8.12 \\ (1.63$\times$)} & \makecell{8.83 \\(1.60$\times$)} & \makecell{10.41 \\ (1.47$\times$)} &  \makecell{11.44 \\ (1.42$\times$)} & \makecell{11.9 \\ (1.36$\times$)} \\
    \bottomrule
    \end{tabular}
    }
    \label{tab:latency2}
\end{center}
\end{table}

\subsection{Qualitative Analysis}
In this section, we analyze how \tool enhances the token acceptance rate, and which tokens the draft model acquires across varying query distributions.

{\bf High-frequency tokens precision and recall.} In the experiment on the Spider dataset, (LLaMA-160M draft, Vicuna-7B target). 
We identify the top 100 tokens most frequently generated by the target model, which account for 72.2\% of all appearances, 
following a power-law distribution. Figure~\ref{fig:freq-acc} shows a marked improvement in both accuracy and recall
of these tokens after distillation. 


{\bf Tokens learned across different datasets}
We analyze the top 10 tokens with the most pronounced accuracy and recall improvements across various datasets, focusing on the 100 most frequent tokens to 
understand the draft model's learning trends. As detailed in Table~\ref{tab:tokens}, the improved tokens align with the underlying data distribution. 
For example, in the Spider SQL dataset, tokens like SELECT and WHERE have notably higher acceptance rates post-distillation. 
These patterns highlight the draft model's ability to adapt and predict tokens consistent with the data distribution.





\subsection{\tool and Medusa}
Next, we demonstrate the application of \tool to Medusa. By frequently updating the weights of the linear layer during non-peak hours, \tool can further improve the token acceptance rate on the given dataset and achieve further speedup. 

Medusa~\cite{cai2024medusa} differs from standard speculative decoding in two key ways: (1) Standard speculative decoding proposes a single token for each position, whereas Medusa proposes multiple token candidates for each position. (2) Standard speculative decoding uses a small, independent draft model to propose tokens, while Medusa employs additional heads—linear layers that take the last hidden states as input and output the token probability distribution for each position.

In Table \ref{tab:osd-medusa}, we compare the performance of Medusa-v1 combined with \tool. By fine-tuning the additional heads in Medusa on the Spider dataset, this combination achieves a 2.01$\times$ speedup, compared to the 1.34$\times$ improvement observed with Medusa alone. Notably, when standard speculative decoding with online updates is applied to the Spider dataset, it delivers even greater performance gains, achieving a 2.17$\times$ speedup compared to the 2.01$\times$ enhancement provided by Medusa with \tool.
On the Arena dataset, Medusa-v1 generally performs well on the chat dataset since the extra heads are originally trained on the ShareGPT~\cite{sharegpt} dataset, showcasing a 2.03$\times$ speedup without any updates. In contrast, the combination of standard speculative decoding with \tool yields worse results than vanilla Medusa-v1, with only a 1.51$\times$ speedup. Nevertheless, \tool still manages to enhance the performance of Medusa-v1 by 0.35$\times$. This underscores \tool's potential in further optimizing the tree-style speculative decoding method.

\begin{table}[ht!]
\caption{\tool and Medusa}
\begin{center}
    \resizebox{0.8\linewidth}{!}{
    \begin{tabular}{c|c|c|c}
    \toprule
    Dataset     & Spider & Chatbot Arena & Extra Parameters (B)\\
    \midrule
    Medusa-7B            & 1.34$\times$ & 2.03$\times$ & 0.44 \\
    Medusa-7B + \tool      & 2.01$\times$ & 2.38$\times$ & 0.44 \\
    Draft model + \tool & 2.17$\times$ & 1.51$\times$      & 0.16 \\
    \bottomrule
    \end{tabular}}
    \label{tab:osd-medusa}
\end{center}
\end{table}
\vspace{-10pt}
\section{Conclusion}
Speculative decoding’s efficiently hinges on the draft model’s approximation to the target model. We
introduce an online speculative method that continuously enhances the draft model based on varying
data distributions. Experiments on both synthetic and real data demonstrate that online speculative
decoding swiftly adapts to new data distributions, significantly enhancing token acceptance rate.

\section*{Impact Statement}
This paper presents work whose goal is to advance the field of Machine Learning. There are many potential societal consequences of our work, none of which we feel must be specifically highlighted here.

\section*{Acknowledgments}
Z.J. Deng was supported by NSF of China (No. 62306176), Natural Science Foundation of Shanghai (No. 23ZR1428700), Key R\&D Program of Shandong Province, China (2023CXGC010112), and CCF-Baichuan-Ebtech Foundation Model Fund. This work is supported in part by the National Science Foundation through grants CCF-2238346, IIS-1955488, IIS-2027575, ARO W911NF2110339, ONR N00014-21-1-2724, and DOE award DE-SC0016260, DE-SC0021982.





\nocite{langley00}

\bibliography{example_paper}
\bibliographystyle{icml2024}

\newpage
\appendix
\onecolumn
\section{Appendix}
\subsection{Speculative Decoding Sampling}
\label{appendix:sd}
With $i$ rising from $1$ to $k$, speculative decoding accepts the proposal ${y}_i$ if $u \leq  p(y_i|\vx, {\vy}_{<i}) / q_\vtheta({y}_i|\vx, {\vy}_{<i})$ where $u \sim U[0,1]$; otherwise exits. 
Let $a$ denote the number of accepted tokens, which takes values in $\{0,\dots, k\}$. 
We can sample an additional token ${y}_{a+1}$ from the following distribution 
\begin{equation}
\label{eq:speedup}
\footnotesize
p'(y) =
    \begin{cases}
      p(y|\vx, {\vy}_{<a+1}) & \text{if $a = k$}\\
      \mathrm{norm}(\max(0, \\ \quad p(y|\vx, {\vy}_{<a+1}) - q_\vtheta(y|\vx, {\vy}_{<a+1}))) & \text{otherwise}
    \end{cases}       
\end{equation}
where $\mathrm{norm}(\cdot)$ makes the probabilities over the vocabulary sum to $1$. 

Prior work has shown that the resulting samples $\tilde{\vy} \triangleq \{{y}_1, \dots, y_{a+1}\}$ strictly follow the distribution of the target LLM $p(\cdot|\vx)$~\citep{leviathan2023fast}. 
We concatenate $\tilde{\vy}$ to $\vx$ and repeat the above process until meeting ⟨EOS⟩. 
Each run of the target LLM generates $a+1$ tokens with $a\geq0$. This ensures that at least one new token is generated even in the worst case. 
The generation process can be significantly accelerated if the draft LLM better approximates the target one, particularly $a$ is larger for each target LLM run.

\subsection{Speculative Decoding Speedup}
\begin{figure}[h]       
    \centering
    \includegraphics[width=0.2\linewidth]{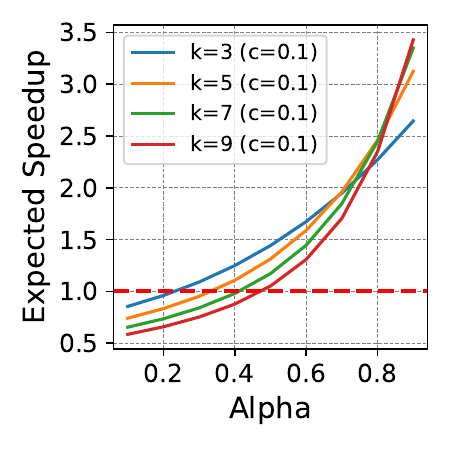}
    \includegraphics[width=0.2\linewidth]{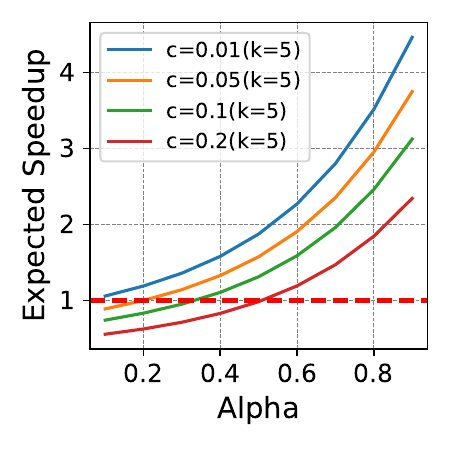}
    \vspace{-10pt}
    \caption{Speculative decoding speedups for different \(\alpha\)s. $c$: time ratio for a single run between the draft and target model. $k$: propose length.  For small \(\alpha\), speculative decoding may slow down inference (speedup \(<1\)). The connection between speedup and \(\alpha\) is superlinear; doubling the acceptance yields over 2$\times$ speedup.}
    \label{fig:analysis-alphas}
\end{figure}

As proved in ~\cite{leviathan2023fast}, compared with standard decoding, the expected improvement factor for offline speculative decoding is \(\frac{1-\alpha^{k+1}}{(1-\alpha)(ck+1)}\).
Let the time taken for a single run of \(M_p\) be \(T\). Define \(c\), the cost coefficient, as the ratio of the time taken for a single run of \(M_q\) to that of \(M_p\).
Each execution of lines 7 to 8 takes \(Tck + T\) and, on average, yields \(\frac{1-\alpha^{k+1}}{1-\alpha}\) tokens.
As a result, the average time to produce one token using speculative decoding is given by \(\frac{(ck+1)(1-\alpha)}{1-\alpha^{k+1}}T\). 
In contrast, the time to produce a single token using standard decoding is \(T\). 
Hence, the wallclock time reduction of offline speculative decoding can be described as \(\frac{1-\alpha^{k+1}}{(1-\alpha)(ck+1)}\).

\subsection{Latency Analysis}
\label{appendix:latency-analysis}
Suppose \tool can improve the token acceptance rate from \( \alpha_1 \) to \( \alpha_2 \) 
and $T$ is the generation time for standard decoding. Based on Equation~\ref{eq:gen_len}, this improvement leads to a decrease in the average generation time for each token, transitioning from \( \frac{(ck+1)(1-\alpha_1)}{1-\alpha_{1}^{k+1}}T \) to \( \frac{(ck+1)(1-\alpha_2)}{1-\alpha_{2}^{k+1}}T \). Consequently, this results in a speedup factor of \( \frac{1-\alpha_2^{k+1}}{1-\alpha_1^{k+1}}\frac{1-\alpha_1}{1-\alpha_2} = \frac{1+\alpha_2+\alpha_2^2+...+\alpha_2^{k}}{1+\alpha_1+\alpha_1^2+...+\alpha_1^k}\) compared to standard speculative decoding.

In the aforementioned analysis, we omitted the additional latency due to updating the smaller model for the following reasons:
(1) As illustrated subsequently, the additional computational cost (FLOPs) from the update remains marginal when juxtaposed with the 
computational demands of running the larger model.
(2) Updates are periodic, during times of moderate request loads, the latency for serving individual requests remains largely unaffected. 
Additionally, given that the update operation for the smaller model is considerably less resource-intensive than inference, 
the associated latency might be seamlessly masked, rendering it virtually imperceptible.
Lastly, the processes of updating and inference can even be executed concurrently on separate devices.

\subsection{Flops Analysis}
\label{appendix:flops}
\emph{The FLOPs required to update the draft model are significantly fewer than those needed for inference on a large model.}
Denote \(L\) as the average length of the generated sequence. 
For each verification, the draft model suggests \(k\) tokens. 
The expected length for a single run of the target LLM, denoted as \(a\), can be calculated using Equation~\ref{eq:gen_len}. 
Therefore, \tool undergoes the verification process \(\frac{L}{a}\) times, with each time verifying $k+1$ tokens.
We use \(F_{qfwd}\) to represent the arithmetic operations required by a singular forward run of the draft model for each token, 
and \(F_{pfwd}\) stands for the FLOPs needed for a single forward run of the target model per token.
Therefore, the computational demand (in FLOPs) for the draft and teacher models to handle one request can be expressed as:
$
\text{FLOPs}(draft)  = \frac{L}{a} \times k \times F_{qfwd},
\text{FLOPs}(target) = \frac{L}{a} \times (k+1) \times F_{pfwd}.
$
Let's consider the FLOPs required to update the student model per token as \(F_{qbwd}\). The cumulative FLOPs necessary to process \(I\) requests is given by:
\[
\frac{LI}{a} \times \left[k \times F_{qfwd} + (k+1) \times F_{pfwd}\right] + I \times L \times F_{qbwd}.
\]
Based on the findings of \cite{kaplan2020scaling}, training is approximately three times costlier than inference. This translates to roughly 6 FLOPs per parameter for training on a single token and 2 FLOPs per parameter for inferring on one token. Thus, we can simplify the total FLOPs expression to:
\begin{equation}
    \frac{LI}{a}\left[(k + 3a) \times F_{qfwd} + (k+1) \times F_{pfwd}\right].
\end{equation}

The proportion of FLOPs needed to run the target model to that of the draft model is given by:
\[
\frac{(k+1)\times F_{pfwd}}{(k+3a)\times F_{qfwd}}.
\]
For the two model pairs evaluated, assuming an average of 5 proposed tokens per run: 
(1) (LLaMA-160M, Vicuna7B) with an average acceptance rate of 0.71, the ratio is approximately \( \frac{(5+1) \times 7B}{(5+3 \times 3) \times 160M} = 18.75 \).
(2) (T5-small 80M, Flan-T5-XL 3B), with an average acceptance rate of 0.76, the ratio is roughly \( \frac{(5+1) \times 3B}{(5+3 \times 4.3) \times 80M} = 12.6 \).

\emph{In practical systems, the FLOPs required for inference are significantly below the machine's capacity.} 
Consider the LMSYS-Chat-1M~\cite{zheng2023lmsyschat1m}. It comprises traces spanning 125 days with 1000,000 requests, averaging less than 2,000 tokens per request (including both prompts and responses).
When serving a 30B model with 8 A100 GPUs, the FLOPs consumed per second can be estimated as (Still, we estimate 2 FLOPs per token per parameter):
\[ \frac{2000 \times 1000,000}{125 \times 24 \times 3600} \times 30 \times 10^9 \times 2 = 5.5 \times 10^9 \text{ FLOPs or 5.5 GFLOPs} \]
On the other hand, 8 A100 GPUs offer a combined capacity of \( 8 \times 312 \text{ TFLOPs} \), and the computational utilization is notably low. 
While Arena (the platform that generates LMSYS-Chat-1M) may not be the most efficient and might lack substantial traffic, it's the only publicly accessible LLM service trace. 
Even after amplifying the load multiple times, based on the above calculations, the computation efficiency remains limited.

\subsection{Bandwidth Analysis}
\label{appendix:bandwidth}
LLM inference is memory bandwidth bound. When the input/output length is short, the memory operations are dominated by loading model parameters from GPU HBM to SRAM.
We analyze the memory loading requirements of different inference techniques below ($batch\_size=1$).
We first introduce the notations used in the analysis. $M$/$m$: The total bytes of the target/draft model. $L$: inference length. 
$a_1$/$a_2$: The expected generation length for a single run of the target LLM of Vanilla speculative decoding(VSD)/OSD. $I$:  the interval to update the draft model.
On a high level, $\frac{L}{a} * M$ represents the bytes required to load the target model, while $L * m$ indicates the bytes needed for 
loading the draft model. For OSD, $m * \frac{L}{I}$ denotes the bytes necessary to load the draft model for updates.

We applied Formula~\ref{eq:gen_len} from our paper to calculate $a_1$, $a_2$, using the token acceptance rates for standard vanilla speculative decoding and OSD on the 
Spider dataset with the LLaMA-160M and Vicuna-7B models as the draft and target models, respectively. 
This resulted in $a_1 = 1.4$ and $a_2 = 3.4$. The memory sizes are $M$ = 14GB for the target model and $m$ = 0.32GB for the draft model. 
For OSD, the draft model is updated every 8 iterations ($I$=8). Using these values, we have estimated the memory loading bytes, presented in the right column.

\begin{table}
\caption{Bandwidth analysis. Original means inference without speculative decoding. VSD, vanilla Speculative Decoding. OSD, online speculative decoding.}
\begin{center}
    \resizebox{0.7\linewidth}{!}{
    \begin{tabular}{c|c|c}
    \toprule
     & Memory Loading Formula & \begin{tabular}[c]{@{}c@{}}Memory Loading in bytes of\\ (LLaMA-160M, Vicuna-7B) pair, $L$=128, $a_1$=1.4, $a_2$=3.4 \end{tabular} \\
     \midrule
     Original & $L * M$ & 1792 GB \\
     VSD & $\frac{L}{a_1} * M + L * m $ & 1320 GB \\
     OSD & $\frac{L}{a_2} * M + L * m + m * \frac{L}{I} $ & 573 GB \\
    \bottomrule
    \end{tabular}}
    \label{tab:bw}
\end{center}
\end{table}

\subsection{Data Mix}
\label{appendix:data-mix}
\begin{figure}      
    \centering
    \includegraphics[width=0.4\linewidth]{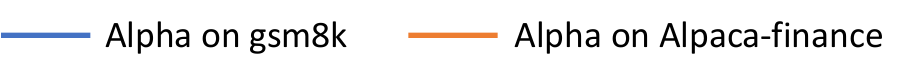} \\
    \includegraphics[width=0.4\linewidth]{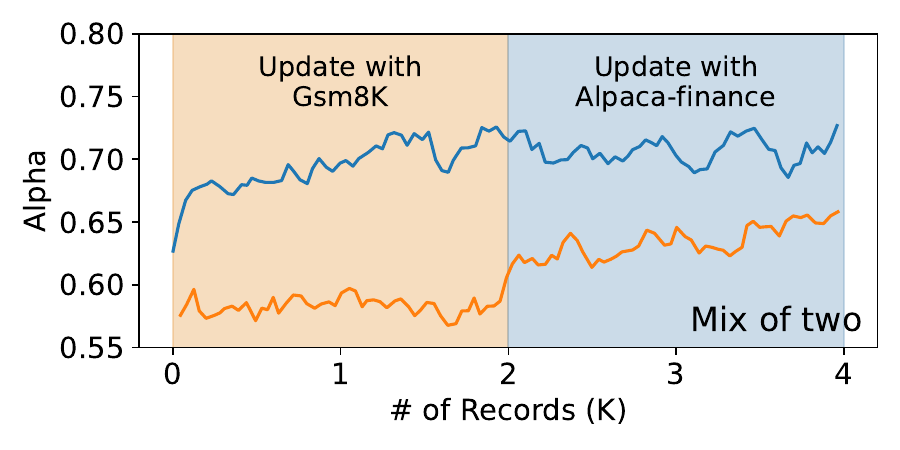}
    \caption{Mix of distributions.}
    \label{fig:mix}
\end{figure}

Moreover, there is a question of whether the draft model, once adapted to the new distribution, might lose its prior knowledge. 
To probe this, we conducted an experiment mixing 2k prompts each from the Gsm8k and Alpaca-finance datasets. 
During online serving, for the initial 2k requests, we only update the model based on data from the Gsm8k dataset. 
For the subsequent half of the requests, we restrict updates solely to data from the Alpaca-finance dataset. 
We then provide the average token acceptance rates for all requests, segmented by their data source (Gsm8k versus Alpaca-finance).
As depicted in Figure~\ref{fig:mix}, the token acceptance rate for Gsm8k increases as the draft model 
is exposed to more data. Conversely, the acceptance rate (\(\alpha\)) for the Alpaca-finance dataset remains consistent. 
This is anticipated since we only update the draft model using Gsm8k data. In the latter half of the dataset, 
the token acceptance rate for the Alpaca-finance dataset also shows an uptrend. Intriguingly, the rate for Gsm8k remains consistent, 
suggesting that the draft model retains its learned knowledge without showing signs of forgetting.

\subsection{Real Workloads}
\label{appendix:arena}
\begin{figure}[t]  
    \centering
    \includegraphics[width=0.3\linewidth]{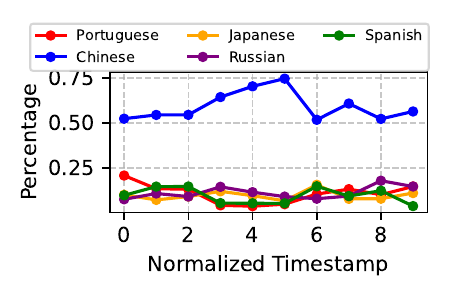}
    \includegraphics[width=0.3\linewidth]{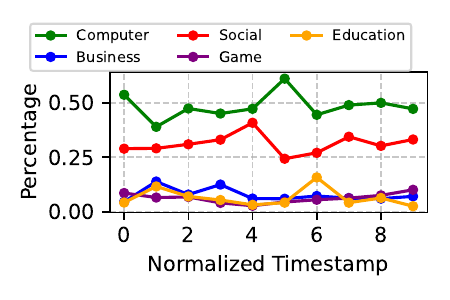}
    \vspace{-20pt}
    \caption{We divided the time into ten equal segments. For each segment, the percentage indicates the share of queries relative to the total within that specific interval. The left image graphically represents the evolution of query distribution among various languages over these periods. In contrast, the right image categorizes these queries by topic, showcasing their topical distribution varying over time.}
    \vspace{-20pt}
    \label{fig:topic-per}
\end{figure}

\textbf{Arena Dataset}
For expedited experimental evaluation, we randomly sample a subset with 10K records
from LMSYS-Chat-1M~\cite{zheng2023lmsyschat1m}, 
a comprehensive real-world LLM conversation dataset. This dataset encompasses interactions with 25 models spanning 
from April to August 2023 and features conversations in over 150 languages. For all experiments, we only pick conversations for Vicuna models.

\subsection{Various Degrees of Distribution Shifts}

In this section, we experimented to evaluate if OSD can maintain a high token acceptance rate amidst varying degrees of distribution shifts. We conduct simulations to replicate both abrupt and gradual distribution changes. In the abrupt shift scenario, we merge the Gsm8k and Spider datasets, each containing 2,000 records, without any transitional phase. Conversely, for gradual distribution shift, we introduce a probabilistic blend of the two datasets. Here, the likelihood of a record originating from the Gsm8k dataset decreases linearly from 100\% to 0\% as we progress from the first to the 4,000th record, while the probability of it coming from the Spider dataset increases correspondingly. This results in a smooth transition from Gsm8k to Spider data.

We measure the token acceptance rate by averaging it over the most recent 100 records. Our findings show a notable dip in the token acceptance rate at the juncture of the two datasets (around the 2,000th record) for the abrupt shift case. However, the acceptance rate quickly recovers. In the case of the gradual shift, the decline in the token acceptance rate is much less discernible, indicating that OSD can indeed effectively adapt to progressive changes in the data distribution by consistently maintaining a high token acceptance rate.

\begin{figure}[h]
    \centering
    \includegraphics[width=0.3\linewidth]{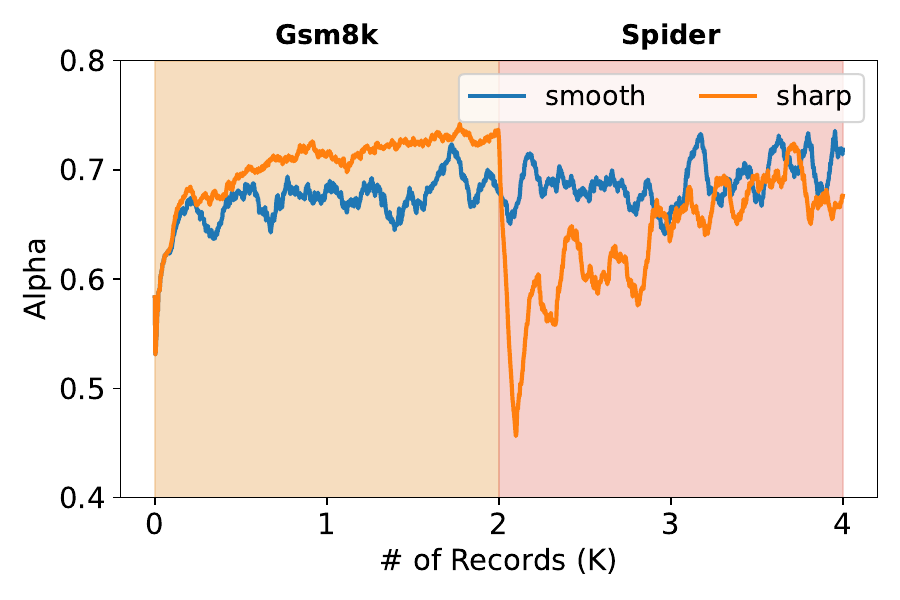}
    \caption{Vary degrees of distribution shift.}
    \label{fig:dis-shift-appendix}
    \vspace{-20pt}
\end{figure}

\begin{table}[ht!]
\caption{Top 15 tokens with the most recall/precision improvement across datasets. We ignore \_ before tokens, which represents space in the LLaMA tokenizer.}
\label{tab:tokens}
\begin{center}
\begin{scriptsize}
\begin{tabular}{p{2.4cm}|p{2.8cm}|p{2.8cm}|p{2.8cm}|p{2.8cm}}
\toprule
{\bf Dataset} & {\bf Spider} & {\bf Gsm8k} & {\bf Alpaca-Finance} & {\bf Code-Python} \\
\midrule
{\bf Tokens with the greatest precision increase}
&
AV, SELECT, first, ⟨EOS⟩, template, SUM, G, COUNT, \textbackslash n, city, WHERE, ';, (, IST, id
&
⟨EOS⟩, \textgreater\textgreater, +, To, \textless\textless, this, =, \%, know, are, We, calculate, be, The, have
&
1, Here, (, :, provide, depends, However, goals, amount, 3, there, The, \textbackslash n, personal, will
&
''', (, Here, python, ', how, doc, snippet, import, based, \{, Python, This, :, you \\
\hline
{\bf Tokens with the greatest recall increase}
&
SELECT, *, FROM, (, IST, *), \textbackslash n, COUNT, G, first, WHERE, ⟨EOS⟩, IN, ;, MAX, ';
&
start, \textgreater\textgreater, \textless\textless, +, find, how, we, =, fore, To, so, \textbackslash, ⟨EOS⟩, then, let
&
general, 1, several, This, depends, Here, provide, However, goals, over, (, If, amount, it, can
&
Here, This, snippet, ''', ', how, python, (, takes, Python, you, doc, an, import, def \\
\bottomrule
\end{tabular}
\end{scriptsize}
\end{center}
\end{table}

\begin{table}[ht!]
\caption{Measured execution time/speedup and theoretical execution time/speedup. Original means inference without speculative decoding. The numbers are measured on a single A100-80G with batch size = 1 and draft token length $k = 8$.}
\begin{center}
    \resizebox{0.99\linewidth}{!}{
    \begin{tabular}{p{6cm}|p{2cm}|p{3cm}|p{3cm}|p{3cm}|p{3cm}|p{3cm}}
    \toprule
     & Original &  OSD, $\alpha=0.5$ & OSD, $\alpha=0.6$ &  OSD, $\alpha=0.7$ & OSD, $\alpha=0.8$ & OSD, $\alpha=0.9$ \\
      \midrule 
    \multicolumn{7}{c}{Vicuna-33B + TinyLLaMA-1.1B $\left( c=0.08 \right)$} \\
     \midrule
     Measured time in ms/token (speedup) & 51.09 & 39.90 (1.28 $\times$) & 35.48 (1.44 $\times$) & 30.96 (1.65 $\times$) & 25.42 (2.01 $\times$) & 19.43 (2.63 $\times$) \\
      \hline
     Theoretical time in ms/token (speedup)  & 51.09 & 39.00 (1.31 $\times) $ & 32.12 (1.59 $\times$) & 26.07 (1.96 $\times$) & 20.77 (2.46 $\times$) & 16.38 (3.12 $\times$) \\ 
     \midrule
     \multicolumn{7}{c}{Vicuna-7B + LLaMA-160M $\left( c=0.13 \right)$} \\
     \midrule
     Measured time in ms/token (speedup) & 13.21  & 13.85 (0.95$\times$) & 11.39 (1.17$\times$) & 10.20 (1.3$\times$)  &  7.84 ($1.68\times$) & 5.17 (2.55$\times$) \\
      \hline
     Theoretical time in ms/token (speedup) & 13.21 & 13.48 (0.98 $\times$) & 10.92 (1.21 $\times$) & 
     8.41 (1.57$\times$) & 6.23 (2.12$\times$) & 4.40 (3.00$\times$) \\ 
    \bottomrule
    \end{tabular}}
    \label{tab:latency-theoretical-measured}
\end{center}
\end{table}

\subsection{Theoretical and Measured Latency}
\label{sec:theoretical-measured-latency-analysis}
In this section, we compare the theoretical speedup calculated with Formula~\ref{eq:speedup} and measured speedup.
As shown in Table~\ref{tab:latency-theoretical-measured}, the observed speedup closely aligns with the theoretical expectations. Slow sampling can be the reason for the discrepancies. Speculative decoding necessitates additional sampling steps, as the draft model generates preliminary tokens. For optimal performance, the sampling process must be expedited. Moreover, to attain significant speedup, the execution time ratio (denoted as $c$) between the draft and target models should be minimized. However, in practical implementations, the overall execution time for the draft model is disproportionately affected by kernel launch overheads and Python-related delays, resulting in slower-than-anticipated performance. 

\subsubsection{Generation Length and Speedup}
\begin{table}[ht!]
    \centering
    \resizebox{0.6\linewidth}{!}{
    \begin{tabular}{c|c|c|c|c|c}
         \toprule
         Generation length & $\alpha=0.5$ & $\alpha=0.6$ & $\alpha=0.7$ & $\alpha=0.8$ & $\alpha=0.9$ \\
         \midrule
         16                & 1.22         & 1.37         &  1.52        & 2.01         & 2.40          \\
         32                & 1.23         & 1.41         &  1.52        & 2.05         & 2.42          \\
         64                & 1.23         & 1.41         &  1.53        & 2.05         & 2.43          \\
         128               & 1.24         & 1.42         &  1.53        & 2.05  
               & 2.43          \\
         256               & 1.25         & 1.45         &  1.54        & 2.07   
               & 2.46          \\
         512               & 1.25         & 1.48         &  1.61        & 2.12   
               & 2.57          \\
         1024              & 1.27         & 1.52         &  1.66        & 2.17  
               & 2.56          \\
         2048              & 1.29         & 1.55         &  1.73        & 2.24   
               & 2.70          \\  
        \bottomrule
    \end{tabular}
    }
    \caption{Generation length and measure speedup across different token acceptance rates.}
    \label{tab:gen-len-speedup}
\end{table}
In this section, we test the effect of generation length on measured speedup, we conduct the following experiments in llama.cpp. With our implementation of \tool, we fix the batch size at 1, prefix length at 512, draft token length at 5. We vary the generation length from 16 to 2048, and evaluate speedup at different token acceptance rate alpha. The teacher and student models used are Vicuna-7B-v1.5 and Llama-160M respectively.

From the Table~\ref{tab:gen-len-speedup}, we see that the measured speedup does increase at each token acceptance rate as generation length increases from 16 to 2048. This is because, at greater generation lengths, each decoding step executed by the teacher model becomes increasingly more memory-intensive. Employing a student model can thus reduce the time ratio $c$ in equation (1), representing a single run comparison between the teacher and student models, thereby achieving a greater speedup. However, the data presented in the table reveals that the impact of extending generation length on speedup is not as pronounced as the effect of enhancing the token acceptance rate. This suggests that improvements in token acceptance rate are more effective in achieving significant speedups than simply increasing generation length.

\subsubsection{Draft Model Execution Time}
To verify that when the draft model is small, updating the draft model (including the forward and backward pass) is much less expensive than inference on the target model, we measure the execution time of updating the draft model and the large model inference time (prompt\_len=64, generation\_len=128) on a single A100-80G with Huggingface Transformer library~\cite{hftransformer}.
Assuming we update the draft model every eight iterations (k=8), the forward execution time is 24.96/0.12 = 208 times longer than the time spent updating the draft model for the (160M, 7B) pair, and 55.12/0.67 = 82.3 times longer for the (1.1B, 33B) pair.

\begin{table}[ht!]
    \centering
    \resizebox{0.99\linewidth}{!}{
    \begin{tabular}{c|c|c|c|c|c}
         \toprule
         Model Pairs                     & \begin{tabular}[c]{@{}c@{}}Draft  forward +\\ backward + update time (s)\end{tabular} &
         Target  forward time (s) of & Target forward time (s) of k iterations (k=8)         & 
         Draft model update time (s) every k iterations (k=8) \\
         \midrule
         (160M, 7B)  & 0.12 & 3.12 & 24.96 & 0.12 \\
         (1.1B, 33B) & 0.67 & 6.89 & 55.12 & 0.67 \\
        \bottomrule
    \end{tabular}
    }
    \caption{Generation length and measure speedup across different token acceptance rates.}
    \label{tab:gen-len-speedup}
\end{table}



\end{document}